%% file: mia.tex
\documentclass[journal]{IEEEtran}

\usepackage{framed,multirow}
\usepackage{amssymb}
\usepackage{latexsym}
\usepackage{url}
\usepackage{amsmath,amssymb,amsfonts}
\usepackage{algorithmic}
\usepackage{graphicx}
\usepackage{textcomp}
\usepackage{times}
\usepackage{epsfig}
\usepackage{soul}
\usepackage{mathtools}
\usepackage{dsfont}
\usepackage{tabularx}
\usepackage{booktabs}
\usepackage[table,xcdraw]{xcolor}

\definecolor{mypink1}{rgb}{0.858, 0.188, 0.478}
\definecolor{newcolor}{rgb}{.8,.349,.1}
\newcommand{\Angie}[1]{{\color{black} #1}}

\begin{document}

\title{Contrastive Registration for \\ Unsupervised Medical Image Segmentation}

\author{Lihao Liu, 
        Angelica I Aviles-Rivero, and Carola-Bibiane Schönlieb
\thanks{L. Liu., A.I. Aviles-Rivero and C. Schönlieb are with the Department of Applied Mathematics and Theoretical Physics, University of Cambridge. Cambridge CB3 0WA, UK.  Corresponding author: ll610@cam.ac.uk.}}

\maketitle

\input{section/abstract}

\input{section/introduction}

\input{section/related_work}
\input{section/methodology}

\input{section/experiments}

\input{section/conclusion}

\bibliographystyle{IEEEtran.bst}
\bibliography{refs}

\end{document}

%% file: section/abstract.tex
\begin{abstract}
Medical image segmentation is an important task in medical imaging, as it serves as the first step for clinical diagnosis and treatment planning.
Whilst major success has been reported using deep learning supervised techniques, they assume a large and well-representative labelled set. This is a strong assumption in the medical domain where annotations are expensive, time-consuming, and inherent to human bias. 
To address this problem, unsupervised segmentation techniques have been proposed in the literature. Yet, none of the existing unsupervised segmentation techniques reach accuracies that come even near to the state of the art of supervised segmentation methods.
In this work, we present a novel optimisation model framed in a new CNN-based contrastive registration architecture for unsupervised medical image segmentation \Angie{called CLMorph}. The core idea of our approach is to exploit image-level registration and feature-level contrastive learning, to perform registration-based segmentation. Firstly, we propose an architecture to capture the image-to-image transformation mapping via registration for unsupervised medical image segmentation. Secondly, we embed a contrastive learning mechanism in the registration architecture to enhance the discriminative capacity of the network at the feature level. \Angie{We show that our proposed CLMorph technique mitigates the major drawbacks of existing unsupervised techniques.} We demonstrate, through numerical and visual experiments, that our technique substantially outperforms the current state-of-the-art unsupervised segmentation methods on two major medical image datasets. 
%


\end{abstract}
%

%% file: section/introduction.tex
\section{Introduction}

Medical image segmentation is the task of partitioning an image into multiple regions, which ideally reflect qualities such as well-defined structures guided by boundaries in the image domain. 
This task has been successfully applied in a range of medical applications using data coming from different parts of human anatomy including the heart~\cite{chen2020deep,liu2021rethinking}, lungs~\cite{hu2001automatic} and brain~\cite{de2015deep}.
Medical image segmentation is a relevant task; as it is the first step for several clinical applications such as tumor localisation and neuropsychiatric disorder diagnosis. 
%

The body of literature has reported several successful techniques for medical image segmentation, in which major  progress has been reported using fully-supervised convolutional neural networks (CNNs)~\cite{ronneberger2015u,isensee2021nnu,chen2018voxresnet,chen2019med3d} or partially-supervised CNN methods~\cite{zhao2019data,wang2020lt,portela2014semi,cui2019semi}.
%
These techniques rely on learning prior mapping information from pair-wise data (mapping from medical images to their manual segmentation) to achieve astonishing performance at the same level as, and sometimes outperforming, radiologists.
However, a major constraint of these methods is the assumption of having a well-representative set of annotations for training, which is not always possible in the medical domain. Moreover, obtaining such annotations is time-consuming, expensive, and requires expert knowledge. 
%
%
To deal with these constraints,  a body of research has explored unsupervised segmentation techniquess~\cite{alfano1997unsupervised,lee1998unsupervised,dalca2018anatomical,dalca2019unsupervised}, in which no annotations are needed. 
%

Notably, in recent years unsupervised techniques have become a great focus of attention as they do not require segmentation labels. 
In this context, clustering algorithms are often applied to perform unsupervised segmentation by grouping the image content with similar intensities~\cite{ng2006medical,li2011integrating,jose2014brain,tian2017deepcluster}.
%
However, these methods are still limited performance-wise; as it is difficult to learn any image-to-mask transformation mapping when relying solely on images.
To further embed the transformation mapping information within an unsupervised architecture for better medical image segmentation performance, \Angie{one feasible solution is to cast the segmentation task as an unsupervised deformable registration problem~\cite{krebs2019learning,zhao2019unsupervised,de2019deep,zhang2018inverse,dalca2019unsupervised,dalca2018unsupervised,balakrishnan2019voxelmorph,balakrishnan2018unsupervised,mok2020large,zhao2019recursive,heinrich2019closing}.} 
The goal of this perspective is to find an optimal image-to-image transformation mapping to align a set of images in one coordinate system.

In this work, we aim to find an optimal image-to-image transformation mapping $z$ between an unaligned image $x$ and a reference image $y$, which is usually formulated by a metric function $\phi_z$. 
Benefiting from similar morphological attributes between the medical image and its corresponding segmentation mask (such as the shape of different organs), one can also transfer the segmentation mask of the reference image $y_{seg}$ back to the coordinate system of the unaligned image. 
This process is with the purpose of obtaining the segmentation mask of the unaligned image $x_{seg}$ using the optimal mapping information $z$.
Hence, when the image is correctly registered, the segmentation mask is automatically obtained, which we called \textit{registration-based segmentation} (aka atlas-based segmentation).

Following this philosophy, we develop an unsupervised segmentation model based on a registration architecture.
A central observation of existing registration methods is that they only focus on capturing the mapping information on the image level, but fail to enhance the feature-level representation.
\Angie{Most recently, unsupervised feature representation learning has demonstrated promising results. In particular, contrastive-based models, such as SimCLR~\cite{chen2020simple} and BYOL~\cite{grill2020bootstrap}, are reaching performance comparable to those produced by supervised techniques for different tasks.
The main idea is that by contrasting images to others, the differences between images are easily remembered within a network (i.e., learning distinctiveness); thus making the learned feature more robust to discriminate images with different labels.}
Therefore, to further improve the feature-level learning by contrasting the unaligned image $x$ to the reference image $y$, we propose to embed the contrast feature learning in the registration architecture to extract feature maps with richer information, producing better unsupervised segmentation results.

\Angie{In this paper, we present a novel \textbf{C}ontrastive \textbf{L}earning registration architecture based on Voxel\textbf{Morph} for unsupervised medical image segmentation, which we named named \textbf{CLMorph}.
Our proposed CLMorph is a simple yet effective unsupervised segmentation model.}
Unlike existing techniques, our approach combines image-level registration and feature-level contrastive representation learning.
More specifically, our technique works as follows.
We first propose two weight-shared feature encoders, where CNN features are extracted from the unaligned images and reference images, respectively.
Then, by contrasting the extracted CNN features from the unaligned images and reference images, one can obtain CNN features with richer information.
Moreover, we use one single decoder to capture the mapping information $z$ from the contrasted CNN features.
Finally, we use the spatial transform network of~\cite{jaderberg2015spatial} along with the captured $z$ to align the segmentation mask, of the reference image to the coordinate system of unaligned images, to obtain the segmentation mask in an unsupervised manner.
Our main contributions are: 

\begin{itemize}
    \item We propose a simple yet effective contrastive registration architecture for unsupervised medical image segmentation, \Angie{in which we highlight the combination of image-level registration architecture and feature-level contrastive learning, which we called CLMorph}. 
    To the best of our knowledge, this is the first work that embeds contrastive learning in a registration architecture for unsupervised segmentation. 
  
    \item We evaluate our technique using a range of numerical and visual results on two major benchmark datasets. We show that, by casting the unsupervised segmentation task via registration 
    with feature-level contrast, we can largely improve the unsupervised segmentation performance and reduce the performance gap between supervised and unsupervised segmentation techniques.
    
    \item We demonstrate that our contrastive registration architecture, as a by-product, can also lead to a better registration performance than the state-of-the-art unsupervised registration techniques.
\end{itemize}

%% file: section/related_work.tex
\section{Related Work}
The body of literature has reported impressive results for image segmentation. 
In this section, we review the existing techniques in turn.

\begin{figure*}[th]
    \centering
    \includegraphics[width=\textwidth]{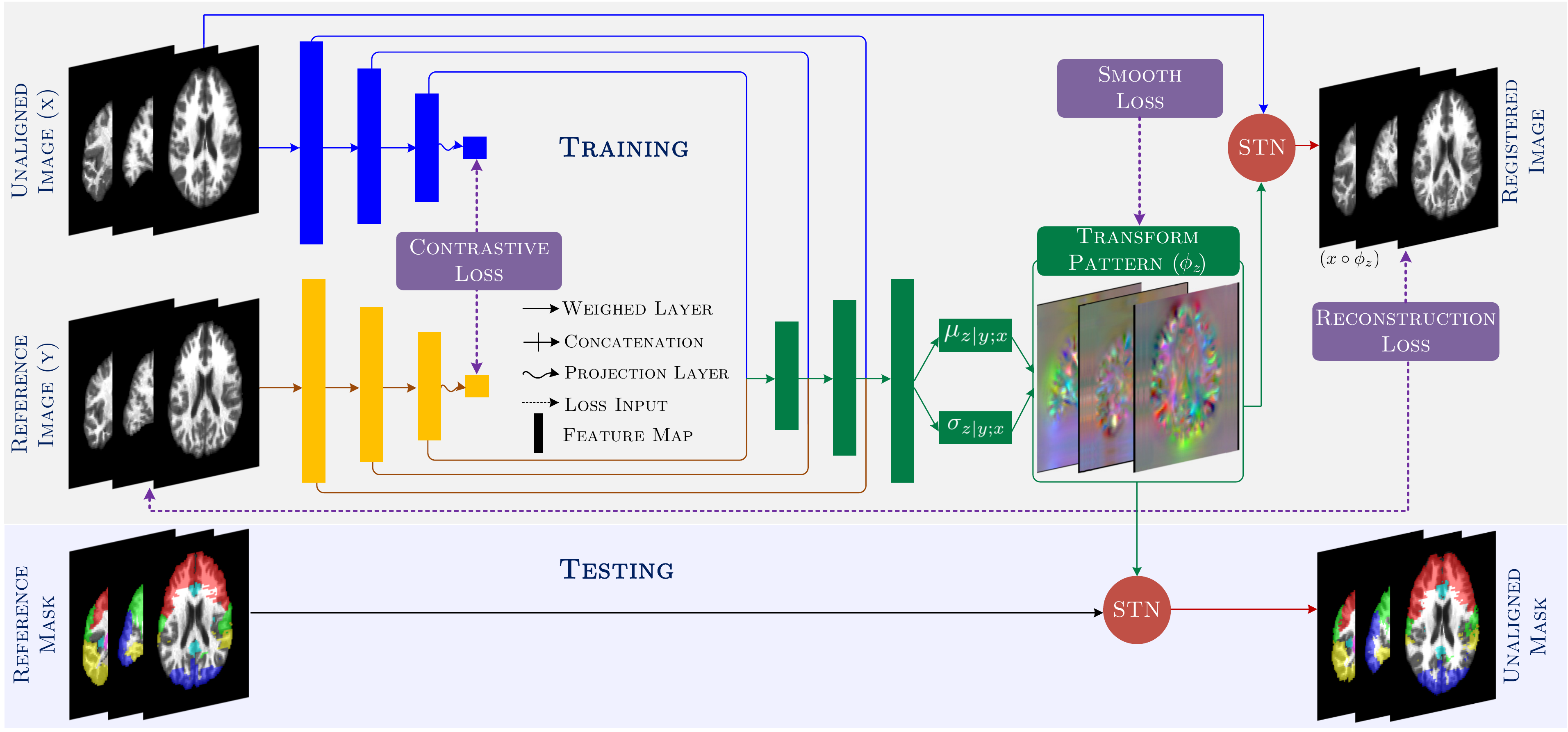}
    \centering
    \caption{\Angie{The overall workflow of the proposed network architecture: CLMorph. It takes two images as input for the two CNNs, which generates two sets of feature maps (see blue and yellow blocks). Two projection layers are applied to the feature maps to produce two vector presentations of the input images. Based on these vector presentations, a contrastive loss is employed.  It then recursively concatenates and enlarges the contrasted CNN feature maps, from the two CNNs, using a single decoder until the transformation mapping is obtained. Based on the transformation mapping, the reconstruction and smooth loss are adopted to ensure that the registration is well-performed. After the above training process is completed, we adopt the learned transformation mapping to transfer the reference mask to obtain the segmentation mask of the unaligned image.} 
    }
    \label{fig:network}
\end{figure*}

%
\subsection{Supervised Medical Image Segmentation}
In this section, we revise exiting techniques that heavily rely on annotations (i.e. supervised techniques), and the close related paradigm of one-shot learning where labels are still needed for the technique to work.

\textbf{Fully-supervised medical image segmentation.} Medical image segmentation has been extensively investigated in the literature, in which supervised methods have been most successful.
Early works learn to segment the different organs based on hand-crafted features including thresholding~\cite{suzuki1991automatic}, statistical model~\cite{fischl2002whole} and Bayesian model~\cite{yang2007automatic}.
%
%
%
%
%
%
However, a major drawback is that they are not capable of capturing high-level semantic information. 
Thereby, they tend to fail in segmenting those organs accurately.

More recently, convolutional neural networks (CNN) based methods, which train with annotations to capture the high-level semantic information, have demonstrated remarkable performance beyond radiologist execution, such as U-Net~\cite{ronneberger2015u}, nnU-Net~\cite{isensee2021nnu}, VoxResNet~\cite{chen2018voxresnet}, and MedicalNet~\cite{chen2019med3d}.
%
Although effective in biomedical image segmentation, these supervised methods rely heavily on a well-representative set of annotations. This is a strong assumption in the medical domain 
as annotations are expensive to obtain (in many medical tasks there is a need for at least two readers), highly uncertain, and require expert knowledge.
Hence, when applying the trained model to another dataset such approaches often fail to segment correctly.
%

%
%

\textbf{One-shot medical image segmentation.} Recently, the community has moved towards the perspective of developing techniques that require a  tiny labelled set. 
For example, the principles of one-shot learning have been reported in the literature for medical image segmentation. For example, the authors of~\cite{zhao2019data}  proposed DataAug, which uses the learned spatial and intensity transformation to synthesize labelled images for one-shot medical image segmentation. Another work using this philosophy is LT-Net~\cite{wang2020lt}. In that work, the authors embedded a cycle consistency loss into a registration architecture to perform one-shot medical image segmentation.
However, \textit{in the training process, one still needs to provide a  set of labels that are well-representative for the task at hand. }
An alternative to those sets of techniques is unsupervised learning which has recently been a great focus of attention in the medical imaging community. This paradigm is the focus of this work and existing techniques are discussed next.

\subsection{Unsupervised Medical Image Segmentation}
\textbf{Clustering.} For unsupervised segmentation, clustering algorithms~\cite{ng2006medical,li2011integrating,jose2014brain,tian2017deepcluster} have been extensively explored. The idea is to divide the image into different groups of pixels/voxels according to the similarity of image intensities (i.e., each pixel value). 
%
%
%
%
Besides clustering techniques, works based on Autoencoders~\cite{baur2020scale} and GANs~\cite{song2020unsupervised} have also been explored for unsupervised medical image segmentation.
However, due to the lack of segmentation masks, which determines if it is impossible to learn any image-to-mask mapping information, these algorithms are still inefficient for the task of segmentation. 

\textbf{Registration-based segmentation.} To enable the mapping information in the training process, \Angie{another feasible solution is to cast the segmentation task an unsupervised deformable registration process~\cite{krebs2019learning,zhao2019unsupervised,de2019deep,zhang2018inverse,dalca2019unsupervised,dalca2018unsupervised,balakrishnan2019voxelmorph,balakrishnan2018unsupervised,mok2020large,zhao2019recursive,heinrich2019closing}.}
Instead of seeking the image-to-mask transformation mapping, the goal of unsupervised registration methods is to find an optimal image-to-image transformation mapping.
Specifically, given an unaligned image $x$ and a reference image $y$, the main goal of unsupervised registration methods is to calculate the latent variable $z$, which contains the image-to-image transformation mapping information~\cite{balakrishnan2019voxelmorph,balakrishnan2018unsupervised}.
Benefited from the morphological similarity between the medical image and its corresponding segmentation mask, unsupervised registration architecture can transfer the segmentation mask of the reference image $y_{seg}$ back to the coordinate system of the unaligned image to obtain the segmentation mask of the unaligned image $x_{seg}$, using the optimal transformation mapping $z$.  
Hence, as long as the image is correctly registered, the segmentation mask is automatically obtained, which we call registered segmentation.

\textbf{Contrastive learning.} The current unsupervised registration methods only focus on capturing the mapping information from the image level and ignore the feature-level representation learning.
\Angie{To further enhance the feature representation learning without annotations, a body of researchers has demonstrated promising results using contrastive representation learning including Momentum Contrast (MoCo)~\cite{he2020momentum}, SimCLR~\cite{chen2020simple}, Contrastive Multiview Coding (CMC)~\cite{tian2019contrastive}, and BYOL~\cite{grill2020bootstrap}.
The main idea of contrastive representation learning is to maximise the differences between images from different groups as well as to maximise the agreements between images and their different augmented views, such as SimCLR~\cite{chen2020simple}.
Another research line relies on teacher-student learning mechanisms to learn from each other without group difference (no negative pairs), such as BYOL~\cite{grill2020bootstrap}.
Both directions are effective for learning features more robust to discriminate images from different groups. We show these findings in our experiments.
}

Current contrastive learning methods mainly focus on improving the discriminating capacity of CNN-based models on image classification tasks. Likewise, other techniques directly employ contrastive learning methods for the downstream segmentation tasks in a supervised setting~\cite{zhao2020contrastive} or semi-supervised manner~\cite{zhao2020contrastive,chaitanya2020contrastive}.
To further transfer the robust feature-level learning of the contrastive learning mechanism to the downstream tasks (i.e., segmentation), we propose to embed the contrast feature learning in the registration architecture to extract feature maps with richer information for unsupervised segmentation. 
\textit{To our best knowledge, this is the first work that embeds contrastive learning in a registration architecture for unsupervised segmentation}

%% file: section/methodology.tex
\section{Proposed Technique}
This section contains two key parts, the description of: i) the registration architecture which employs a CNN-based probabilistic model to calculate image-to-image transformation mapping $z$, and ii) our new full optimisation model that is composed of a reconstruction and a smooth loss and combined with a contrastive learning mechanism.

\subsection{{Our Technique Overview}}
In this section, we describe the workflow of our technique, in which we highlight their main components that are described in detail in the following subsections.

We display the overall workflow of our unsupervised segmentation technique in Figure~\ref{fig:network}.
Our technique takes as input two 3D images, the unaligned image $x$ and the reference image $y$, which are fed into a siamese (two-symmetric) weight-shared 3D CNNs to extract the highly semantic feature maps from the unaligned and reference images -- these parts are illustrated in blue and yellow colours in Figure~\ref{fig:network}. 
%
We then employ a contrastive loss on the projection of the two extracted CNN feature maps, which forces the network to contrast the difference between the two CNN feature maps.
The two symmetric weight-shared CNNs can generate more robust CNN feature maps via back-propagating the contrastive loss during training.
%

Based on the contrasted feature maps of the two CNNs, we use a single decoder to integrate all the feature maps, with different resolutions of the two CNNs, and estimate the transformation mapping $z$ -- see the green part in Figure~\ref{fig:network}.
We first concatenate the feature maps with the same resolutions of the two CNNs.
We then adopt a decoder to recursively use the feature maps, with high-level semantic information (low resolutions) to refine the feature maps with low-level detailed information (high resolutions), until we obtain a feature map with the same resolution as the input images. 
We introduce a probabilistic model to estimate the optimal transformation mapping $z$, which is based on the recovered image-resolution feature map.
%
After training and finding the optimal $z$, we obtain the segmentation output by aligning the segmentation mask of the reference images to the coordinate system of the unaligned image. We do this using the optimal $z$ via a spatial transform network~\cite{jaderberg2015spatial}.
%
%
%
\subsection{Transformation Mapping Estimation}
\label{Transformation_Mapping_Estimation}
Given a pair images called unaligned image $x: \mathcal{X} \rightarrow \Omega$  and a reference image $y: \mathcal{X} \rightarrow \Omega$,  where $ \mathcal{X} = [w]\times[h]\times[d] \subset \mathbb{Z}^3$ is the input domain and $\Omega$ is the value data domain (i.e., gray scale), we seek to determine an optimal transformation mapping $z$, which parametrises a spatial transformation function denoted as $\psi_{z}$, such that the transformed unaligned image, ${x\circ\phi_{z}}$, is aligned with $y$.

To compute the transformation mapping, we estimate $z$ by maximising  the posterior registration probability $p(z|y;x)$ given  $x$ and $y$ -- that is, to estimate the central tendency of the posterior probability (maximum a posteriori estimation MAP). However, solving the partition function is intractable and cannot be solved analytically. To address this problem, one can use variational inference and approximate the solution through an optimisation problem over variational parameters. Following this principle, we adopt a CNN-based variational approach to compute $p(z|y;x)$. We first introduce an approximate posterior probability $q_{\psi}(z|y;x)$ which we assume is normally distributed. To measure the similarity between these two distributions, a divergence  $D(q_{\psi}(z|y;x) || p(z|y;x))$ measure can be applied, e.g.,~\cite{amari2009alpha,stein1972bound,minka2005divergence}. We use the most commonly used divergence: the Kullback-Leibler (KL) divergence~\cite{kullback1951information,cover1999elements}. With this purpose, we seek to minimise the KL divergence from $q_{\psi}(z|y;x)$  to $p(z|y;x)$ which expression reads:

\Angie{
\begin{equation} 
\begin{split} \label{Eq:KL-loss}
     \psi^* & = \min_{\psi}\mathrm{KL}[q_{\psi}(z|y;x) \ || \ p(z|y;x)] \\
     &=  \min_{\psi}\mathbb{E}_{q}[log \ q_{\psi}(z|y;x) \ - log \ p(z|y;x)] \\
     &=  \min_{\psi}\mathbb{E}_{q}[log \ q_{\psi}(z|y;x) \ - log \ \frac{p(z,y;x)}{p(y;x)}]  \\
    &=  \min_{\psi}\mathbb{E}_{q}[log \ q_{\psi}(z|y;x) \ - log \ p(z,y;x) + log \ p(y;x)]  \\
    &=  \min_{\psi}\mathbb{E}_{q}[log \ q_{\psi}(z|y;x) \ - log \ p(z,y;x)] + log \ p(y;x)\\
    &=  \min_{\psi}\mathrm{KL}[q_{\psi}(z|y;x) \ || \ p(z)] - \mathbb{E}_{q}[log \ p(y|z;x)] + const, 
\end{split}
\end{equation}
where $\mathbb{E}_{q}[log \ p(y|z;x)]$ is the \textit{evidence}, $\mathrm{KL}[q_{\psi}(z|y;x)$ is \textit{the evidence lower bound (ELBO)}, and $const$ is a normalisation constant.} $q_{\psi}(z|y;x)$ comes from a multivariate normal distribution $\mathcal{N}$:
\begin{equation} \label{Eq:app-loss}
    q_{\psi}(z|y;x)=\mathcal{N}(z;\mu_{z|y;x}, \sigma_{z|y;x}^2) \ ,
\end{equation}

\noindent
where $\mu_{z|y;x}$ and $\sigma_{z|y;x}^2$ denote the mean and variance of the distribution respectively, which can be directly learned through the convolutional layers; as shown in Figure~\ref{fig:network}.
%
Whilst $p(z)$ and $p(y|z;x)$ follow the multivariate normal distribution, which are modelled as:

\begin{equation}
    p(z) = \mathcal{N}(z;0,\sigma_{z}^2) \ , 
\label{eq:Ndistria}
\end{equation}
\begin{equation}
    p(y|z;x) = \mathcal{N}(y;x \circ \phi_{z}, \sigma^2) \ , 
\label{eq:Ndistrib}
\end{equation}

\noindent
where $\sigma_{z}$ is the variance (a diagonal matrix) of this distribution and $x \circ \phi_{z}$ is the noisy observed registered image in which $\sigma^2$ is the variance of the noisy term.
We remark that whilst several works in the literature have addressed the problem of image registration from a probabilistic perspective, for example, the works of that~\cite{krebs2019learning,dalca2019unsupervised}, we emphasise that our framework goes beyond a solely probabilistic principle, and in fact, our novel function is based on other principles such as contrastive term, smooth and reconstruction terms.  They are explained in the next subsection. 

\subsection{Full Optimisation Model}
In this section, we detail our optimisation model which is composed of a reconstruction and smooth loss, and a contrastive mechanism.
%

\textbf{Reconstruction \& Smooth loss.}
According to the derivation from ~\eqref{Eq:KL-loss}, there are two terms to be optimised.
The first term is the KL divergence between the approximate posterior probability $q_{\psi}(z|y;x)$ and the prior probability $p(z)$, and the second term is the expected log-likelihood $E_{q}[log \ p(y|z;x)]$. 
Based on our assumptions from \eqref{Eq:app-loss}-\eqref{eq:Ndistrib}, the derivation is written as:

\begin{equation}
\begin{split}
     \mathcal{L}_{cs}(\psi;x,y) =& \mathrm{KL}[q_{\psi}(z|y;x) \ || \ p(z)] - \mathbb{E}_{q}[log \ p(y|z;x)]\\
    = & \frac{1}{2}[tr(\sigma_{z|y;x}^2) + || \mu_{z|y;x} ||^2 - log \  det(\sigma_{z|y;x}^2)] \\
    & + \frac{1}{2 \sigma^2} || y - x \circ \phi_{z} ||^2.
    \label{eq:registration_losses}
\end{split}
\end{equation}
\Angie{For sake of clarity in the notation, we decouple $\mathcal{L}_{cs}$ to detail the terms. The first term (the second line) is a close form of $KL[q_{\psi}(z|y;x)$.
It enforces the posterior $q_{\psi}(z|y;x)$ to be close to the prior $p(z)$. We called this term the smooth loss:

\begin{equation}
    \mathcal{L}_{smooth} = \frac{1}{2}[tr(\sigma_{z|y;x}^2) + || \mu_{z|y;x} ||^2 - log \  det(\sigma_{z|y;x}^2)],
\end{equation}
where the $log \  det(\sigma_{z|y;x}^2)$ is a Jacobian determinant which spatially smooths the means.}
The second term (the third line) is the expected log-likelihood $E_{q}[log \ p(y|z;x)]$. It enforces the registered image $x \circ \phi_{Z}$ to be similar to reference image $y$, which we refer as our reconstruction loss:

\begin{equation}
    \mathcal{L}_{recon} = \frac{1}{2 \sigma^2} || y - x \circ \phi_{z} ||^2
\end{equation}

\textbf{Contrastive loss.}
\label{Contrastive_Loss}
Contrastive learning is a learning paradigm that seeks to learn distinctiveness. It aims to maximise agreements between images and their augmented views, and from different groups via a contrastive loss in a latent space. In our work, we follow a four components principle for the contrastive learning process, which is described next.

The first component in our contrastive learning process is the 3D images.  The image contents are basically the same including the number of brain structures and the relative locations of each structure.
They are mainly different in structure sizes.
Therefore, we view the unaligned and the reference images as images sampled from different augmented views.
As our second component, we set the two CNN encoders as weight-shared. This with the purpose of ensuring that the CNN-based encoder can extract unified CNN features from the unaligned and reference images.
For the third component, we adopt a fully-connected layer, as the projection layer, to map the CNN features to a latent space where contrastive loss is applied.
Finally, our fourth component is our contrastive loss following the standard definition presented in~\cite{hadsell2006dimensionality,chen2020simple}.

%


From our four-component process, we can now formalise the loss we used in our framework. It is based on a contrastive loss~\cite{hadsell2006dimensionality,chen2020simple,he2020momentum} which is a function whose value is low when the image is similar to its augmented sample and dissimilar to all other samples.
Formally, given a set of images $I$, we view our unaligned image $x$ and reference image $y$ as an augmented image pair, and any other images in $I$ as negative samples.
Moreover, we denote $sim(u, v) = \frac{u^\mathrm{T}v}{||u||\cdot||v||}$ as the cosine similarity between $u$ and $v$.
We then define the contrastive loss function as:\\
\begin{equation}
\begin{split} \label{loss_contrast}
    \mathcal{L}_{contrast} = - log \frac{\exp{(sim(f(x), f(y))/\tau)}}{\sum\limits_{i\in I} \mathds{1}_{i\not= x} \exp{(sim(f(x), f(i))/\tau)}}
\end{split} 
\end{equation}
where $f()$ denotes the CNN encoder, $f(x)$ and $f(y)$ are the generated features from the CNN encoder, and $\mathds{1}_{i\not= x} \in \{0,1\}$  is an indicator, which values 1 only when $i\not= x$. We also define $\tau$ as a temperature hyperparameter~\cite{wu2018unsupervised}.
%

%
\textbf{Optimisation model.}
Our unsupervised framework is composed of three loss functions. 
The first two ones are directly derived from the optimisation model described in \eqref{eq:registration_losses}, in which the image-to-image reconstruction loss and the transformation mapping (deformation field) smooth loss are introduced.
Whilst the third loss is the contrastive loss as in~\eqref{loss_contrast} which forces the network to contrast the difference between unaligned images and reference images.
The total loss is formulated as:

\begin{equation}
\begin{split} \label{total_loss}
    &\mathcal{L}_{total} = \mathcal{L}_{recon} + \alpha\mathcal{L}_{smooth} +  \beta\mathcal{L}_{contrast} \\
    &= \frac{1}{2 \sigma^2} || y - x \circ \phi_{z} ||^2 +  \alpha\bigg(\frac{1}{2}[tr(\sigma_{z|y;x}^2) + || \mu_{z|y;x} ||^2\\
    &- log det(\sigma_{z|y;x}^2)]\bigg) +  \beta\bigg(-log \frac{\exp{(sim(f(x), f(y))/\tau)}}{\sum\limits_{i\in I} \mathds{1}_{i\not= x} \exp{(sim(f(x), f(i))/\tau)}}\bigg),
\end{split}
\end{equation}

\noindent
where $\alpha$ and $\beta$ are the hyper-parameters balancing $\mathcal{L}_{recon}$, $\mathcal{L}_{smooth}$ and $\mathcal{L}_{contrast}$, which we empirically set as 1 and 0.01, respectively. 


%% file: section/experiments.tex
%
\section{Experimental Results}
In this section, we describe the range of experiments that we conducted to validate our proposed technique.

\subsection{Dataset Description \& Evaluation Protocol}

\textbf{Benchmark datasets.} We evaluate our technique using two benchmarking datasets: the LONI Probabilistic Brain Atlas (LPBA40) dataset~\cite{shattuck2008construction} and the MindBoggle101 dataset~\cite{klein2012101}. %
The characteristics are detailed next.

The LPBA40 dataset~\cite{shattuck2008construction} is composed of a series of maps from the regions of the brain.
It contains 40 T1-weighted 3D brain images, from 40 human healthy volunteers, of size  181$\times$217$\times$181 with a uniform space of 1$\times$1$\times$1 mm$^3$. 
The 3D brain volumes were manually segmented to identify 56 structures. 
Whilst the MindBoggle101 dataset~\cite{klein2012101} is composed of a collection of 101 T1-weighted 3D brain MRIs from several public datasets: OASIS-TRT-20, NKI-TRT-20, NKI-RS-22, MMRR-21, and Extra-18.
It contains 101 skull-stripped T1-weighted 3D brain MRI volumes, from healthy subjects, of size 182$\times$218$\times$182 that are evenly spaced by 1$\times$1$\times$1 mm$^3$.
From this dataset, we use 62 MRI images from OASIS-TRT-20, NKI-TRT-20, and NKI-RS-22, since they are already wrapped to MNI152 space and have manual segmentation masks (50 anatomical labels).

\Angie{
\textbf{Evaluation metrics.} To evaluate our technique against the state-of-the-art unsupervised brain image segmentation, we use three widely-used metrics: the Dice similarity coefficient~\cite{dice1945measures}, Hausdorff distance (HD)~\cite{huttenlocher1993comparing}, and average symmetric surface distance (ASSD)~\cite{yeghiazaryan2018family}.
Dice measures the region-based similarity between the predicted segmentation and the ground truth. The higher the Dice is, the better the model performs. Whilst HD and ASSD measure the longest distance (a.k.a. largest difference) and average surface distance (a.k.a. average boundary difference) between the predicted segmentation and the ground truth, respectively. The lower the HD and ASSD is, the better the model performs. The detailed formulation can be found in~\cite{liu2020psi}
}

\input{table/table_1}

\begin{figure}[ht]
    \centering
    \includegraphics[width=\linewidth]{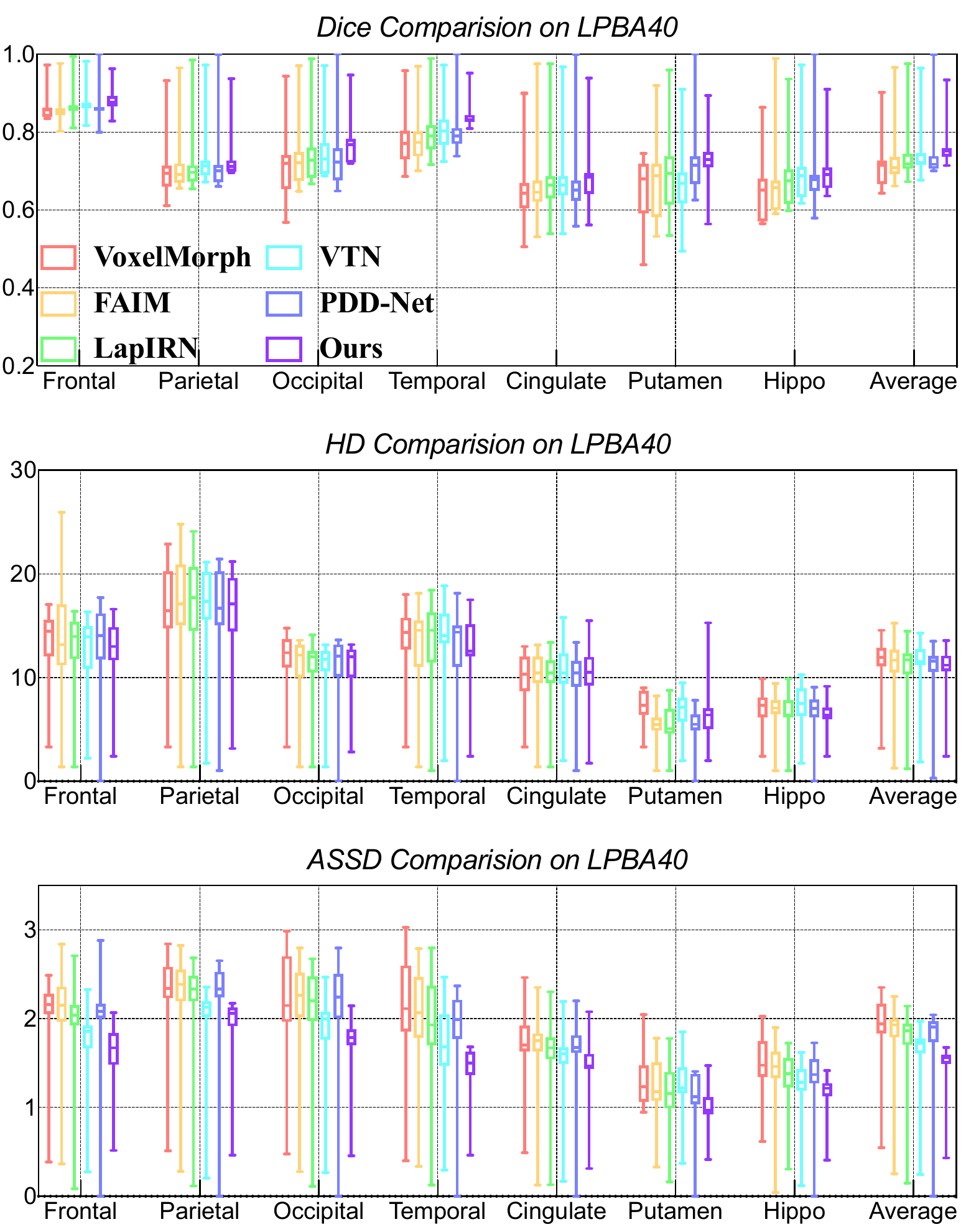}
    \caption{\Angie{Numerical comparisons between our framework and existing SOTA techniques on the LPBA40 dataset. The results are reported in terms of the Dice, HD, and ASSD metrics reflecting a per region, of the brain, evaluation.
    }}
    \label{fig:per_region_lpba40}
\end{figure}
\begin{figure}[ht]
    \centering
    \includegraphics[width=\linewidth]{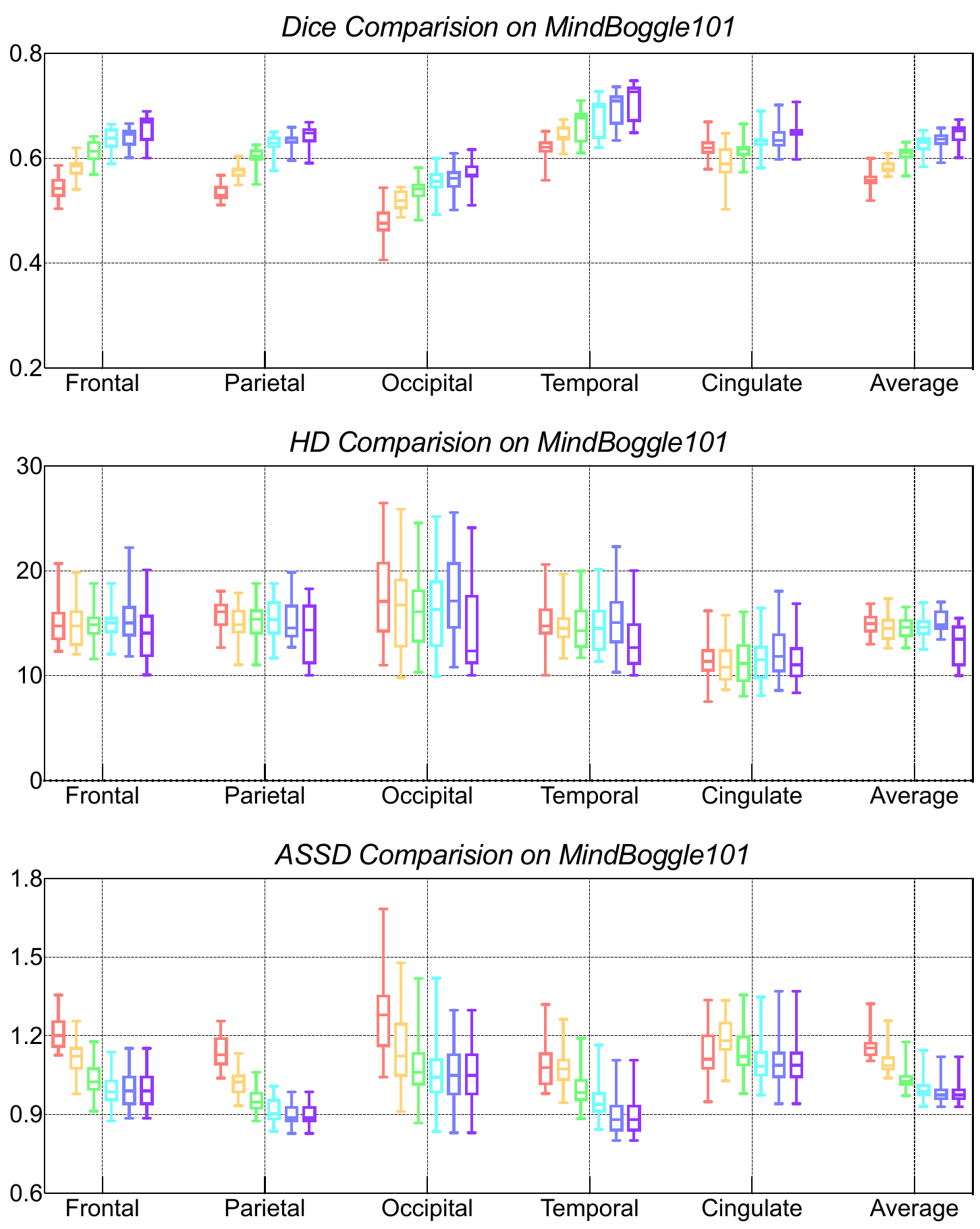}
    \caption{\Angie{Comparisons on the MindBoggle101 dataset. We compare our technique against existing SOTA methods. The comparisons displayed are in terms of the Dice, HD, and ASSD metrics  for each region on the brain.}}
    \label{fig:per_region_mindboggle101}
\end{figure}

\subsection{Implementation Details \& Running Scheme}
In this section, we provide the implementation details and the training \& testing scheme that we followed to produce the reported results.

\begin{figure*}[t]
    \centering
    \includegraphics[width=1\textwidth]{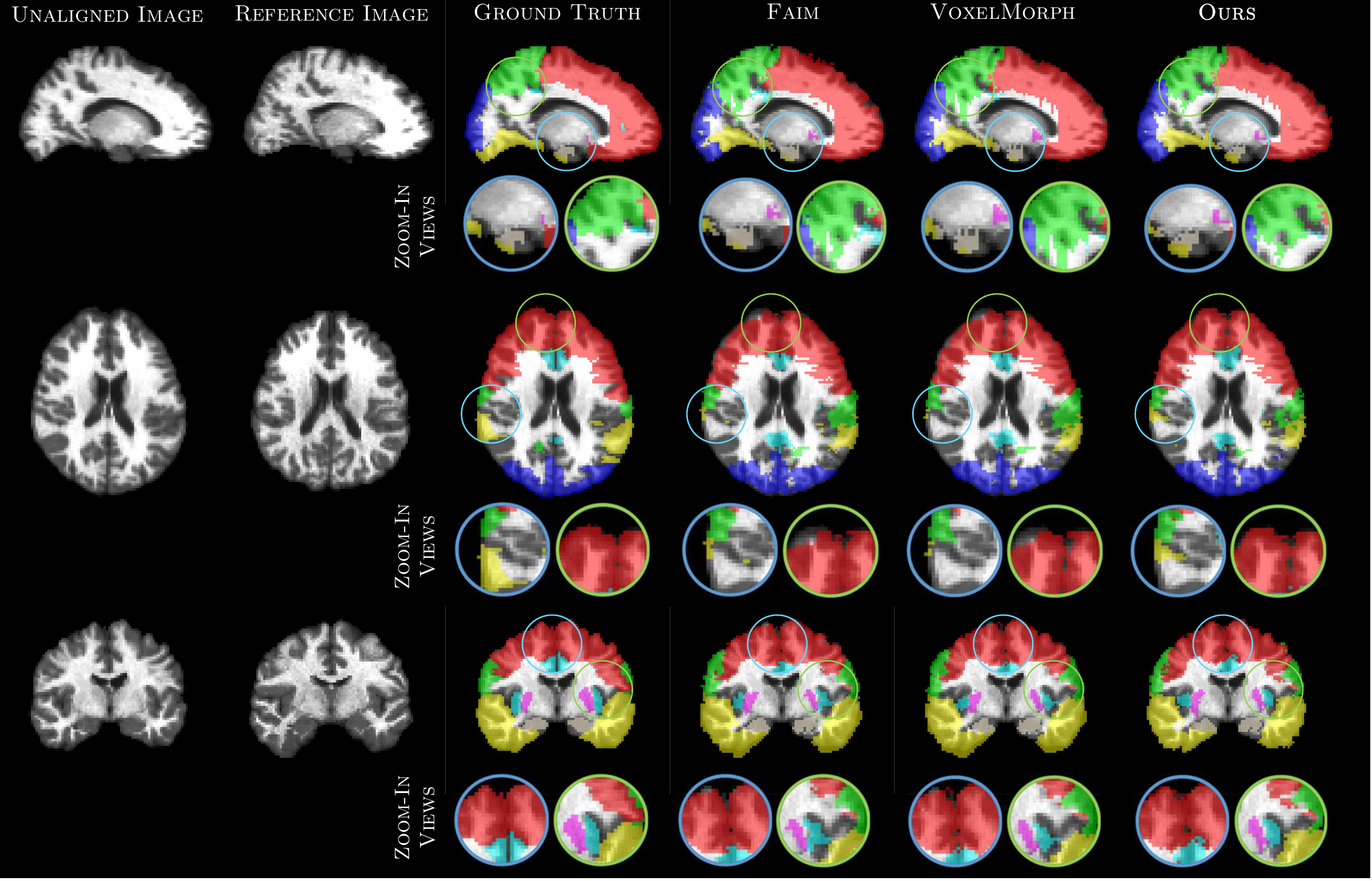}
    \centering
    \caption{Visual comparisons between our technique and unsupervised SOTA techniques for segmentation. The rows show the three views from the 3D data. The third column displays the ground truth whilst the last three samples segmentation from our technique and the compared ones. Zoom-in views are displayed to highlight interesting regions where our technique performs better than the other models.
    }
    \label{fig:segmentation}
\end{figure*}

\textbf{Implementation details.} 
In our experiments,  we follow the training and testing splitting setting from~\cite{liu2019probabilistic} for the LPBA40 dataset, \Angie{where the first 30 images are used as training datasets and the last 10 images are used as testing datasets.}
We group the 56 structures into seven large regions such that we can show the segmentation results more intuitively. 
Moreover and for the MindBoggle101 dataset,  we follow the protocol from~\cite{kuang2019faim} and use 42 images from NKI-TRT-20 and NKI-RS-22 for training and 20 images from OASIS-TRT-20 for testing.
For evaluation purposes, we also group the 50 small regions into five larger regions of interest.
\Angie{As for the reference image, we select the image that is the most similar to the anatomical average as the atlas.
In particular, we selected image \#30 as the reference image for the LPBA40 dataset, and image \#39 for the Mindboggle101.}
Our proposed technique has been implemented in Pytorch~\cite{paszke2019pytorch}.
%

%
We follow the standard pre-processing protocol to normalise the images to have zero mean and unit variance.
We randomly select two images as the pair-wise input data (unaligned image and reference image).
Moreover, we enrich our input data by applying three different transformations (data augmentation) in the following order: random flips in the $y$ coordinate, random rotation with an angle of fewer than 10 degrees, and random crops of size 160$\times$192$\times$160.

\begin{figure*}[th]
    \centering
    \includegraphics[width=1\textwidth]{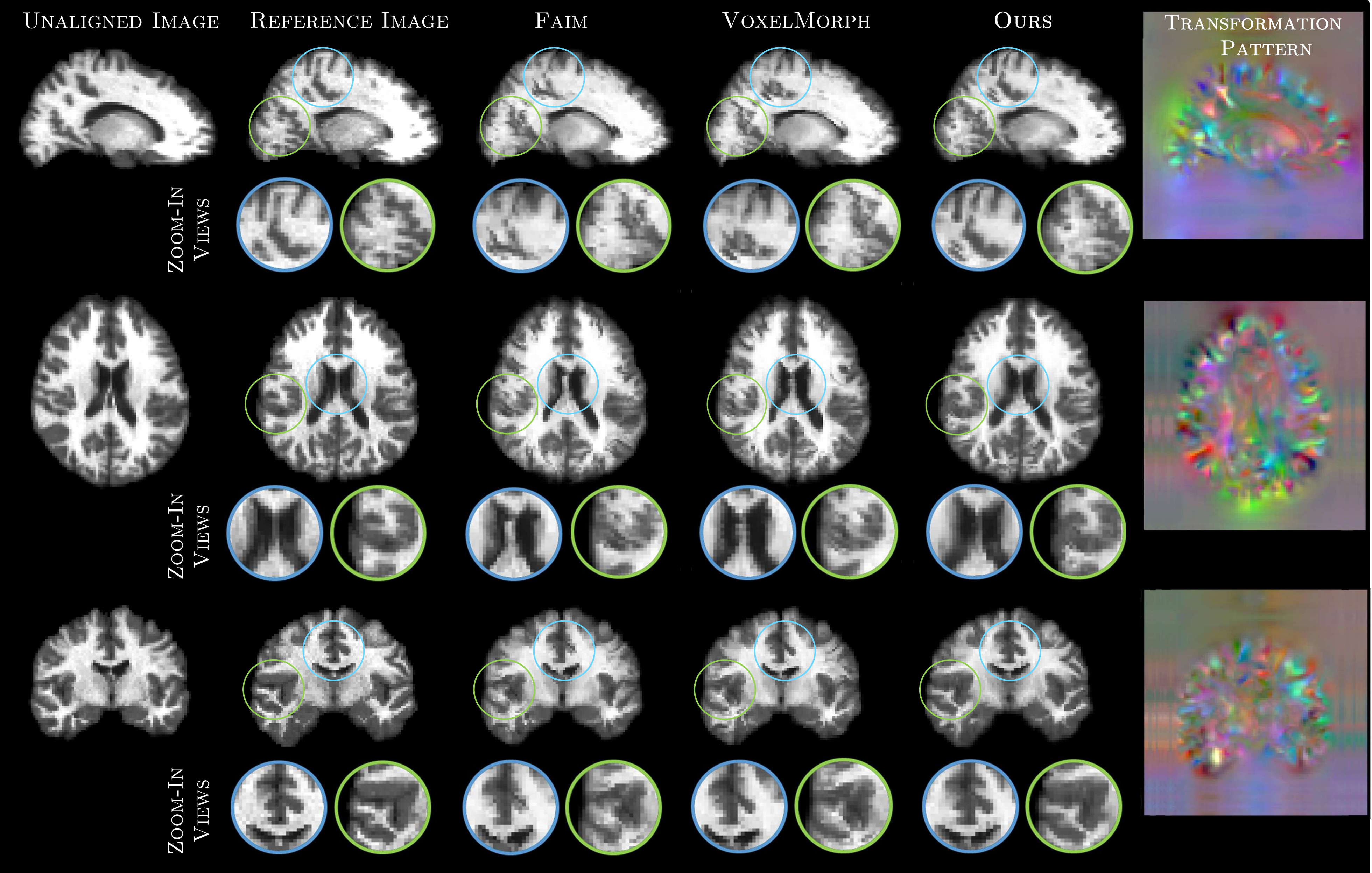}
    \centering
    \caption{Visual comparison of our technique and the SOTA techniques for image registration. The rows display the x, y, and z views from the 3D medical data. The columns display outputs samples from FAIM, VoxelMorph, and our technique. The zoom-in views show interesting structures that clearly show the improvement in terms of preserving the brain structures and fine details.  
    The last column (deformation field) presents the transformation mapping $z$ between unaligned image and reference image produced by our method.}
    \label{fig:registration}
\end{figure*}

\textbf{Training scheme.} 
In our training scheme, we initialise the parameters of all convolutional layers following the initialisation protocol of that~\cite{he2015delving}. 
We also initialise the parameters of all batch normalisation layers using random variables drawn from Gaussian distributions with zero mean and 0.1 derivation~\cite{krizhevsky2017imagenet}.
Moreover, we use Adam optimiser~\cite{kingma2014adam} with a batch size of 8. 
We set the initial learning rate to $3\times 10^{-3}$ and then decrease it by multiplying $0.1$ every $20$ epochs, and terminate the training process after $200$ epochs.
\Angie{Due to the small number of the dataset, which is a common issue in the 3D medical image area, we did not split the data into a validation set from the training or testing sets for model selection. Hence, we use the above learning rate decay policy to decrease the learning rate to a small value in the last few epochs to ensure the model does not diverge, and save the last epoch model as our final model. This is a standard protocol followed in the area.}
Our technique took $\sim20$ hours to train on a single Tesla P100 GPU.
%

%
\textbf{Testing scheme.} 
After training, we performed unsupervised segmentation based on the learned parameters.
We first fed the unaligned image $x$ and the reference image $y$ into the trained network to calculate the transformation relation $z$.
Then, we used a spatial transform network (STN)~\cite{jaderberg2015spatial}, to align the segmentation mask of the reference image according to the calculated $z$, to obtain the segmentation result of the unaligned image.
On average, our method takes less than $10$ seconds to process one whole MRI image on a single GPU (Tesla P100 GPU). 
\subsection{Results \& Discussion}
In this section, we present the numerical and visual results outlined in previous subsections, and discuss our findings and how they are compared with current existing techniques. 

\input{table/table_2}

\begin{figure}[th]
    \centering
    \includegraphics[width=\linewidth]{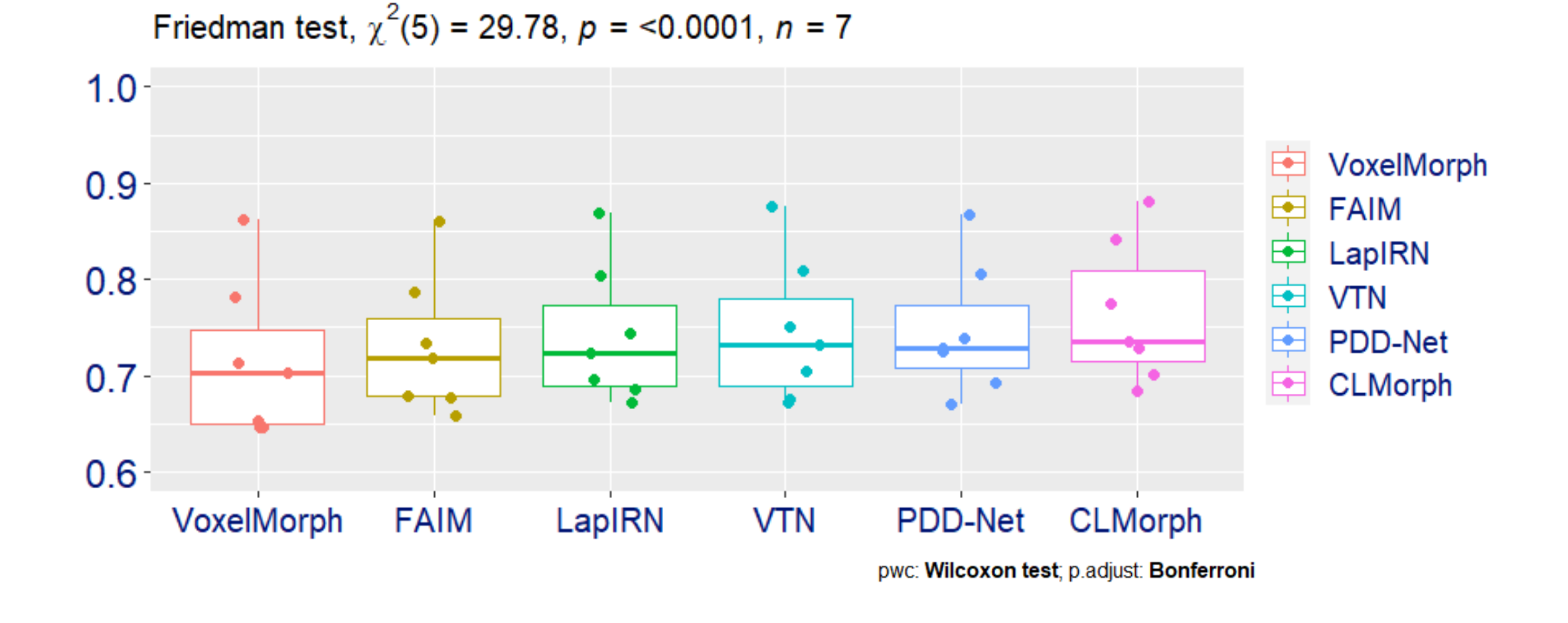}
    \centering
    \caption{\Angie{Statistical analysis. Multiple comparisons are performed followed by  a paired Wilcoxon test and p-values adjusted using the Bonferroni method. Our technique reported significant statistically different among all compared techniques.}}
    \label{fig:p_value}
\end{figure}

\begin{figure}[t]
    \centering
    \includegraphics[width=\linewidth]{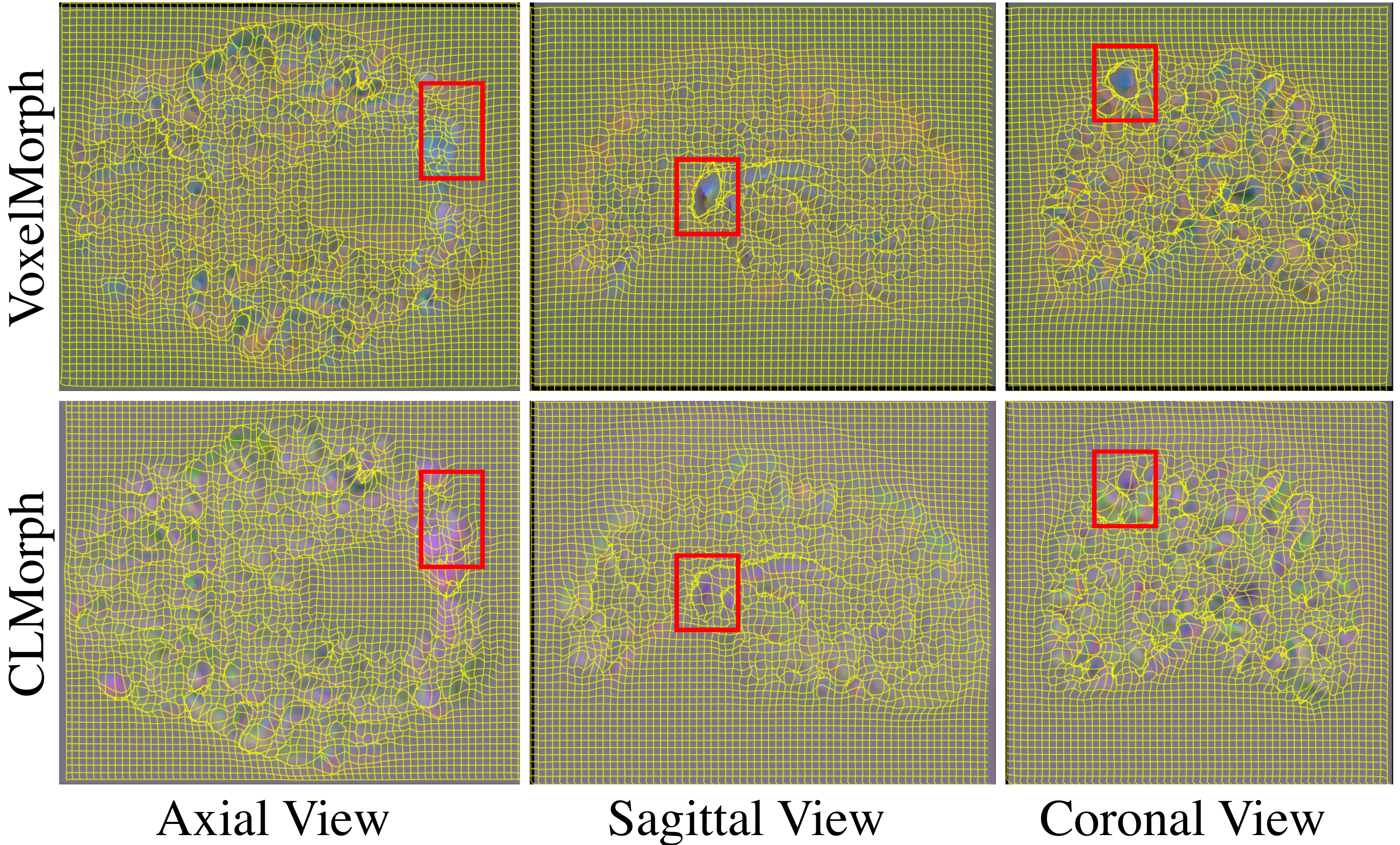}
    \caption{\Angie{Deformation field generated from the non-contrasted feature maps (VoxelMorph) and contrasted feature maps (our CLMorph). The red box highlight the region with large deformation, in which deformation field generated from contrasted feature maps is smoother than the deformation field generated from non-contrasted feature maps.}}
    \label{fig:feature_maps}
\end{figure}

\input{table/table_3}

\textbf{Numerical comparison with the state-of-the-art techniques.}
We begin by   evaluating   our  method  against the state-of-the-art methods unsupervised brain image registration methods on the LPBA40 dataset:  \Angie{UtilzReg~\cite{vialard2012diffeomorphic}, VoxelMorph~\cite{balakrishnan2018unsupervised}, FAIM~\cite{kuang2019faim}, LapIRN~\cite{mok2020large}, VTN~\cite{zhao2019recursive}, and PDD-Net~\cite{heinrich2019closing}}.
\Angie{We remark that we use the registration architecture from other state-of-the-arts methods to predict the transformation mapping $z$, and then adopt the predicted transformation mapping z to perform segmentation as illustrated in the testing stage in Fig.~\ref{fig:network} for image segmentation.}
We report the global results in Table~\ref{new_results} for the \Angie{LPBA40 and MindBoggle101 dataset}, in order to understand the general behaviour and performance of our technique over the SOTA methods. %
\Angie{The displayed numbers are the average of the image metrics across the dataset.}
In a close look at the results, we observe that the compared techniques perform similarly close to each other, whilst our technique outperforms all other techniques \Angie{for all evaluation metrics} by a significant margin. 
This behaviour is consistent on the \Angie{per region results on the two datasets whose results are reported in Fig.~\ref{fig:per_region_lpba40} and Fig.~\ref{fig:per_region_mindboggle101}.}
From these \Angie{boxplots, we can observe that the performance gain of our technique is significant for all regions in both datasets.} 
These comparisons, therefore, show that the combination of the contrastive learning mechanism and the registration architecture can better discriminate brain structures and generate more accurate segmentation outputs.

\begin{figure}[t]
    \centering
    \includegraphics[width=\linewidth]{ 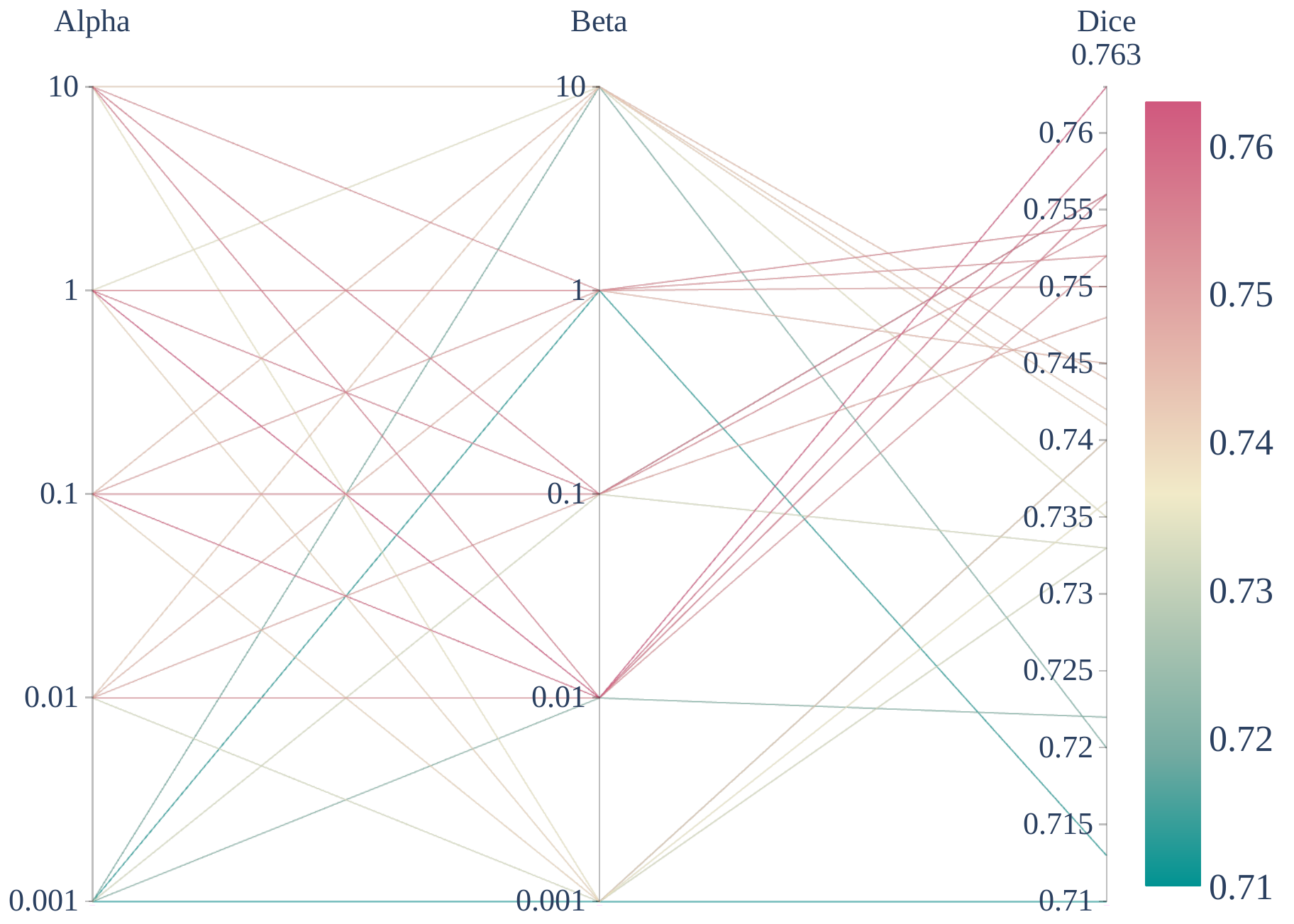}
    \centering
    \caption{\Angie{Hyper-parameter Searching. We display different combination values for $\alpha$ and $\beta$  used in~\eqref{total_loss}. The parameters values are evaluated using the Dice metric. The best combination is $\alpha=1$ and $\beta=0.01$, which gives the highest Dice result of 0.763.}}
    \label{fig:hyperparameter}
\end{figure}

\textbf{Visual comparisons.} To further support our numerical results, we present a set of visual comparisons of a selection of images for our technique and the compared ones.  We start by displaying the unsupervised segmentation outputs in Figure~\ref{fig:segmentation}.
By visual inspection, we observe that the segmentation outputs, generated from FAIM and VoxelMorph, tend to fail to segment correctly several relevant regions of the brain, and they do not adapt correctly to the contour of the brain structure. 
The zoom-in views in Figure~\ref{fig:segmentation} highlight these effects -- for example the red region on the brain that the compared techniques fail to correctly segment; and the yellow region that is not well-captured by FAIM and VoxelMorph.
By contrast, our technique was able to perform better in this regard by capturing fine details in both cases. 
Overall, our technique is able to accommodate better with the fine details in the brain regions, producing segmentation closer to the ground truth.

{We also included the visualisation of the deformation field generated from the contrasted feature (w/ contrastive loss) and non-contrasted feature (w/o contrastive loss). 
According to Fig.~\ref{fig:feature_maps}, when the deformation between the moving and fixed images is large, the deformation field generated from the contrasted feature is smoother, because the contrasted feature maps have high alikeness.
Whilst the deformation field generated from non-contrasted features tends to be less smooth because there is a huge difference between the feature maps extracted from the moving and fixed images.}

\Angie{
\textbf{Statistical Analysis.} We support statistically our visual and numerical findings. To do this, we run a Friedman test for multiple comparisons, $\chi^2(5)$=29.78,$p <0.0001$, followed by a Wilcoxon test for pair-wise comparisons. 
As shown in Fig.~\ref{fig:p_value}, the pair-wise test between groups revealed a statistically significant difference metric-wise of our technique against the rest of compared techniques. 
}

\input{table/table_4}

\textbf{By-product registration accuracy}. As a by-product of our methods, we also provide the numerical and the visual registration results.
Based on the learned deformation field, we register the segmentation mask of the unaligned image to the coordinate system of the reference image to get the aligned segmentation mask.
Then compare it with the reference image's mask to get the dice scores; see Table~\ref{tregistration_results}.
We also present visual results displayed in Figure~\ref{fig:registration}. We observe that our technique was able to better preserve the fine details in the registration outputs than the compared techniques. 
Most notable, this can be observed in the zoom-in views. 
For example, one can observe that our technique is able to preserve better the brain structures whilst the compared techniques fail in these terms.
Moreover, we can see that our registration outputs are closer to the reference image displaying less blurry effects whilst keeping fine details.
We now underline the main message of our paper: \textit{our optimisation model fulfills the intended purposes, and at this point in time our technique outperforms the SOTA unsupervised segmentation results.}
%


\Angie{\textbf{Hyperparameter searching.} We also run a set of new experiments for the hyperparameter analysis. Specifically, we set the weight for the smooth loss ($\alpha$) and contrastive loss ( $\beta$) as 0.001, 0.01, 0.1, 1, 10, respectively. As shown in Fig.~\ref{fig:hyperparameter}, the best combination is our current setting (alpha=1 and beta=0.01), which gives the highest dice result of 0.763.}

%
\textbf{Ablation study.}
To demonstrate that each component of our technique fulfills a purpose, we include an ablation study to evaluate their influence on performance. 
We consider our three major components in our model whose results are displayed in  Table~\ref{ablation_study_results}. 
Our ablation study is performed on the two benchmark datasets, in order to understand the general behaviour of our technique.
The results are displayed in terms of the Dice metric and we progressively evaluate our method with different losses combinations.
From the results in Table~\ref{ablation_study_results}, 
we can observe that whilst the contrastive mechanism, in our technique, indeed provides a positive effect in terms of performance, it benefits our carefully designed components. 
In this work, we also pose the question of -- at this point in time, what is the performance gap between supervised and unsupervised segmentation techniques?
To respond to this question, we use as baseline U-Net~\cite{ronneberger2015u}. 
From the results, in Table~\ref{ablation_study_results}, one can observe that our technique opens the door for further investigation in unsupervised techniques, as the performance shows potential towards the one reported by supervised techniques. 

\Angie{\textbf{Additional experiments on a cardiac dataset.} 
We also provide a set of new results using an additional cardiac dataset called ACDC~\cite{bernard2018deep}. The dataset is for medical image registration. It is composed of 150 labelled images. We selected 100 images for training whilst the remaining 50 images for testing. The results are reported in Table~\ref{cardiac_results}.
From the results, we observe that our model outperforms the compared techniques in terms of Dice and ASSD metrics by a large margin. Whilst readily competing against LVM-S3 in terms of HD metric. These results further support the generalisation and performance of our CLMorph technique.}



%% file: table/table_1.tex
\begin{table*}[t]
\centering
\caption{\Angie{Numerical comparison of our technique vs SOTA techniques for the LPBA40 and MindBoggle101 datasets. The numerical values display the average of Dice, HD, and ASSD over all regions. The higher the Dice is the better the model performs. Whilst, the lower the HD and ASSD are the better the model performs. The best performance is denoted in bold font. The per-region results are illustrated in Fig.~\ref{fig:per_region_lpba40} and Fig.~\ref{fig:per_region_mindboggle101}.} \vspace{0.5cm}}
\resizebox{0.88\textwidth}{!}{
\begin{tabular}{ @{\extracolsep{\fill}}cccccccc } \hline \toprule[1pt]
    \multirow{2}{*}{} & \multicolumn{3}{c}{\cellcolor[HTML]{EFEFEF} LPBA40 \textsc{Dataset}} &   & \multicolumn{3}{c}{ \cellcolor[HTML]{EFEFEF} MindBoggle101 \textsc{Dataset}}                          \\ \cmidrule{2-4} \cmidrule{5-8}
                      &  Dice     & HD      &  ASSD &  &  Dice     & HD      &  ASSD                                \\     \midrule[0.8pt] 
    UtilzReg~\cite{vialard2012diffeomorphic}                                                                & 0.665\scriptsize{$\pm$0.06}                 & 14.16\scriptsize{$\pm$2.34}                & 1.906\scriptsize{$\pm$0.09}                 & & 0.439\scriptsize{$\pm$0.17}                 & 18.21\scriptsize{$\pm$3.04}                & 0.972\scriptsize{$\pm$0.09}               \\  
    SyN~\cite{avants2009advanced}                                                                     & 0.701\scriptsize{$\pm$0.04}                 & 13.61\scriptsize{$\pm$2.56}                & 1.986\scriptsize{$\pm$0.08}                 & & 0.543\scriptsize{$\pm$0.16}                 & 16.83\scriptsize{$\pm$3.68}                & 0.926\scriptsize{$\pm$0.07}               \\ \midrule[0.8pt] 
    VoxelMorph~\cite{balakrishnan2019voxelmorph}                                                              & 0.716\scriptsize{$\pm$0.08}                 & 11.38\scriptsize{$\pm$3.46}                & 1.874\scriptsize{$\pm$0.41}                 & & 0.559\scriptsize{$\pm$0.12}                 & 15.03\scriptsize{$\pm$2.92}                & 1.171\scriptsize{$\pm$0.10}               \\ 
    FAIM~\cite{kuang2019faim}                                                                    & 0.729\scriptsize{$\pm$0.07}                 & 11.04\scriptsize{$\pm$4.11}                & 1.809\scriptsize{$\pm$0.40}                 & & 0.583\scriptsize{$\pm$0.04}                 & 14.50\scriptsize{$\pm$2.71}                & 1.108\scriptsize{$\pm$0.08}               \\ 
    LapIRN~\cite{mok2020large}                                                                  & 0.739\scriptsize{$\pm$0.07}                 & 10.77\scriptsize{$\pm$3.85}                & 1.727\scriptsize{$\pm$0.60}                 & & 0.607\scriptsize{$\pm$0.04}                 & 14.54\scriptsize{$\pm$2.58}                & 1.041\scriptsize{$\pm$0.08}               \\                                                         
    VTN~\cite{zhao2019recursive}                                                                     & 0.745\scriptsize{$\pm$0.08}                 & 11.11\scriptsize{$\pm$3.41}                & 1.601\scriptsize{$\pm$0.31}                 & & 0.626\scriptsize{$\pm$0.04}                 & 14.64\scriptsize{$\pm$2.57}                & 1.007\scriptsize{$\pm$0.09}               \\ 
    PDD-Net~\cite{heinrich2019closing}                                                                 & 0.749\scriptsize{$\pm$0.06}                 & 10.75\scriptsize{$\pm$3.77}                & 1.714\scriptsize{$\pm$0.42}                 & & 0.632\scriptsize{$\pm$0.05}                 & 15.21\scriptsize{$\pm$2.92}                & 0.990\scriptsize{$\pm$0.10}               \\ \midrule[0.8pt] 
    CLMorph (Ours)                                                              & \textbf{0.763\scriptsize{$\pm$0.07}}        & \textbf{10.38\scriptsize{$\pm$2.59}}       & \textbf{1.458\scriptsize{$\pm$0.31}}        & & \textbf{0.646\scriptsize{$\pm$0.04}}        & \textbf{12.76\scriptsize{$\pm$2.52}}       & \textbf{0.892\scriptsize{$\pm$0.05}}      \\ \bottomrule[2pt]
    \end{tabular}
}
\label{new_results}
\end{table*}

%% file: table/table_2.tex
\begin{table}[t!]
    \centering
    \caption{\Angie{Numerical results for our registration process. The reported results denote the average Dice score over all regions of the brain. The best performance is in bold.}}
    \label{tregistration_results}
    \resizebox{1\linewidth}{!}{
    \begin{tabular}{lcc} \hline \toprule[1pt]
    &  \cellcolor[HTML]{EFEFEF} LPBA40    &  \cellcolor[HTML]{EFEFEF} MindBoggle101    \\ \midrule[0.8pt]
    UtilzReg~\cite{vialard2012diffeomorphic}                     & 0.665\scriptsize{$\pm$0.06}           & 0.440\scriptsize{$\pm$0.16}           \\
    SyN~\cite{avants2009advanced}                      & 0.700\scriptsize{$\pm$0.04}           & 0.540\scriptsize{$\pm$0.15}           \\ \midrule[0.8pt]
    VoxelMorph~\cite{balakrishnan2019voxelmorph}               & 0.709\scriptsize{$\pm$0.07}           & 0.558\scriptsize{$\pm$0.10}           \\
    FAIM~\cite{kuang2019faim}                       & 0.720\scriptsize{$\pm$0.08}           & 0.566\scriptsize{$\pm$0.09}           \\
    LapIRN~\cite{mok2020large}                   & 0.727\scriptsize{$\pm$0.09}           & 0.608\scriptsize{$\pm$0.08}           \\
    VTN~\cite{zhao2019recursive}                       & 0.722\scriptsize{$\pm$0.06}           & 0.598\scriptsize{$\pm$0.04}           \\
    PDD-Net~\cite{heinrich2019closing}                  & 0.733\scriptsize{$\pm$0.05}           & 0.612\scriptsize{$\pm$0.08}           \\ \midrule[0.8pt]
    CLMorph (Ours)                     & \textbf{0.750\scriptsize{$\pm$0.08}}  & \textbf{0631\scriptsize{$\pm$0.05}}   \\ \midrule[0.8pt]
    \end{tabular} 
    }
    \end{table}

%% file: table/table_3.tex
\begin{table*}[t]
\caption{\Angie{Ablation Study, in terms of the mean Dice, HD, and ASSD metrics, of our proposed technique. The top part displays a numerical comparison of our technique vs fully supervised segmentation. The bottom part displays a comparison of our technique with its different components. The last two rows display our framework under different contrastive principles BYOL~\cite{grill2020bootstrap}.}}
\centering
\resizebox{0.78\linewidth}{!}{
\begin{tabular}{lcccccc} \hline \toprule[1pt]
&\multicolumn{3}{c}{LPBA40}    &\multicolumn{3}{c}{MindBoggle101}    \\ \midrule[0.8pt]
&  Dice     & HD      &  ASSD   &  Dice     & HD      &  ASSD  \\ \midrule[0.8pt] 
&\multicolumn{6}{c}{ \cellcolor[HTML]{EFEFEF} \textsc{ Fully Supervised Baseline}}        \\ \midrule[0.8pt]
U-Net (Upper Bound)       & 0.832    & - & -    &  0.811    & - & -                      \\ \midrule[0.8pt]
&\multicolumn{6}{c}{ \cellcolor[HTML]{EFEFEF} \textsc{Our Unsupervised Technique}}        \\ \midrule[0.8pt]
$\mathcal{L}_{recon}$                                                               & 0.702\scriptsize{$\pm$0.07}                 & 12.82\scriptsize{$\pm$3.33}                & 1.842\scriptsize{$\pm$0.58}                  & 0.552\scriptsize{$\pm$0.20}                           & 15.11\scriptsize{$\pm$2.26}                & 1.210\scriptsize{$\pm$0.13}                       \\
$\mathcal{L}_{recon}$ + $\mathcal{L}_{smooth}$                                      & 0.716\scriptsize{$\pm$0.08}                 & 11.38\scriptsize{$\pm$3.46}                & 1.874\scriptsize{$\pm$0.41}                  & 0.559\scriptsize{$\pm$0.12}                           & 15.03\scriptsize{$\pm$2.92}                & 1.171\scriptsize{$\pm$0.10}                       \\
$\mathcal{L}_{recon}$ + $\mathcal{L}_{contrast}$                                    & 0.751\scriptsize{$\pm$0.10}                 & 10.92\scriptsize{$\pm$2.28}                & 1.801\scriptsize{$\pm$0.47}                  & 0.604\scriptsize{$\pm$0.14}                           & 14.85\scriptsize{$\pm$2.77}                & 0.993\scriptsize{$\pm$0.21}                       \\ 
$\mathcal{L}_{recon}$ + $\mathcal{L}_{smooth}$ + $\mathcal{L}_{contrast}$ (ours)    & \textbf{0.763\scriptsize{$\pm$0.07}}        & \textbf{10.38\scriptsize{$\pm$2.59}}       & \textbf{1.458\scriptsize{$\pm$0.31}}         & 0.646\scriptsize{$\pm$0.04}                           & \textbf{12.76\scriptsize{$\pm$2.52}}       & \textbf{0.892\scriptsize{$\pm$0.05}}    \\ \midrule[0.8pt]
$\mathcal{L}_{recon}$ + $\mathcal{L}_{byol}$                                        & 0.744\scriptsize{$\pm$0.18}                 & 11.41\scriptsize{$\pm$3.29}                & 1.693\scriptsize{$\pm$0.77}                  & 0.619\scriptsize{$\pm$0.13}                           & 14.73\scriptsize{$\pm$2.39}                & 1.153\scriptsize{$\pm$0.89}             \\
$\mathcal{L}_{recon}$ + $\mathcal{L}_{smooth}$ + $\mathcal{L}_{byol}$               & 0.761\scriptsize{$\pm$0.08}                 & 10.41\scriptsize{$\pm$2.12}                & 1.462\scriptsize{$\pm$0.53}                  & \textbf{0.649\scriptsize{$\pm$0.08}}                  & 12.88\scriptsize{$\pm$2.44}                & 0.894\scriptsize{$\pm$0.10}             \\ \midrule[0.8pt]
\end{tabular}
}
\label{ablation_study_results}
\end{table*}

%% file: table/table_4.tex
\begin{table}[th]
\caption{\Angie{Numerical comparison of our technique vs SOTA techniques for the ACDC Cardiac dataset.}}
\resizebox{\linewidth}{!}{
\centering
\begin{tabular}{ lccc }
\hline \toprule[1pt]
    &  Dice     & HD      &  ASSD                  \\ \midrule[0.8pt]    
    SyN~\cite{avants2009advanced}                                                                     & 0.761\scriptsize{$\pm$0.12}                 & 11.9\scriptsize{$\pm$1.79}                  & 1.581\scriptsize{$\pm$0.07}                  \\ 
    VoxelMorph~\cite{balakrishnan2019voxelmorph}                                                              & 0.787\scriptsize{$\pm$0.09}                 & 11.8\scriptsize{$\pm$2.01}                  & 1.591\scriptsize{$\pm$0.05}                  \\ 
    LVM-S3~\cite{krebs2019learning}                                                                  & 0.795\scriptsize{$\pm$0.06}                 & \textbf{9.90\scriptsize{$\pm$2.22}}          & 1.544\scriptsize{$\pm$0.06}                  \\ \midrule[0.8pt] 
    CLMorph (Ours)                                                              & \textbf{0.810}\scriptsize{$\pm$0.05}        & 10.23\scriptsize{$\pm$2.08}                 & \textbf{1.482}\scriptsize{$\pm$0.05}          \\ \bottomrule[2pt]
    \end{tabular}
}
\label{cardiac_results}
\end{table}

%% file: section/conclusion.tex
%
\section{Conclusion}
This paper presents a novel CNN-based registration architecture for unsupervised medical image segmentation.
Firstly, we proposed  to use unsupervised registration-based segmentation by capturing the image-to-image transformation mapping.
%
Secondly, to promote the image- and feature-level learning, for better segmentation results, we embed a contrastive feature learning mechanism into the registration architecture.
Our network can learn to be more discriminative to the different images via contrasting unaligned image and reference images. We show that our carefully designed optimisation model mitigates some major drawbacks of existing unsupervised techniques. 
We demonstrate, through several experiments, that  our technique is able to report state-of-the-art results for unsupervised medical image segmentation. 
Whilst supervised techniques still report better performance than unsupervised ones, in this work, we show the potentials in terms of performance when no labels are available. This is of a great interest particularly in domains such as the medical area where annotations require expert knowledge and are expensive.
%
%

\section*{Acknowledgements}
LL gratefully acknowledges the financial support from a GSK scholarship and a Girton College Graduate Research Fellowship at the University of Cambridge. AIAR gratefully acknowledges the financial support of the CMIH, CCIMI and C2D3 University of Cambridge.   CBS acknowledges support from the Philip Leverhulme Prize, the Royal Society Wolfson Fellowship, the EPSRC grants EP/S026045/1 and EP/T003553/1, EP/N014588/1, EP/T017961/1, the Wellcome Innovator Award RG98755, the Leverhulme Trust project Unveiling the invisible, the European Union Horizon 2020 research and innovation programme under the Marie Skodowska-Curie grant agreement No. 777826 NoMADS, the Cantab Capital Institute for the Mathematics of Information and the Alan Turing Institute. 